\documentclass[fleqn,10pt]{wlscirep}
\usepackage[utf8]{inputenc}
\usepackage[T1]{fontenc}
\usepackage{lineno}
\usepackage{placeins}

\title{fruit-SALAD: A Style Aligned Artwork Dataset to reveal similarity perception in image embeddings}

\author[1,*]{Tillmann Ohm}
\author[2]{Andres Karjus}
\author[1]{Mikhail Tamm}
\author[3]{Maximilian Schich}
\affil[1]{Tallinn University, School of Digital Technologies}
\affil[2]{Tallinn University, School of Humanities}
\affil[3]{Tallinn University, Baltic Film, Media and Arts School}

\affil[*]{corresponding author: Tillmann Ohm (mail@tillmannohm.com)}

\begin{abstract}
The notion of visual similarity is essential for computer vision, and in applications and studies revolving around vector embeddings of images. However, the scarcity of benchmark datasets poses a significant hurdle in exploring how these models perceive similarity. Here we introduce Style Aligned Artwork Datasets (SALADs), and an example of fruit-SALAD with 10,000 images of fruit depictions. This combined semantic category and style benchmark comprises 100 instances each of 10 easy-to-recognize fruit categories, across 10 easy distinguishable styles. Leveraging a systematic pipeline of generative image synthesis, this visually diverse yet balanced benchmark demonstrates salient differences in semantic category and style similarity weights across various computational models, including machine learning models, feature extraction algorithms, and complexity measures, as well as conceptual models for reference. This meticulously designed dataset offers a controlled and balanced platform for the comparative analysis of similarity perception. The SALAD framework allows the comparison of how these models perform semantic category and style recognition task to go beyond the level of anecdotal knowledge, making it robustly quantifiable and qualitatively interpretable.
\end{abstract}

\begin{document}

\flushbottom
\maketitle
\thispagestyle{empty}

\begin{figure}[h!]
\centering
\includegraphics[width=\linewidth]{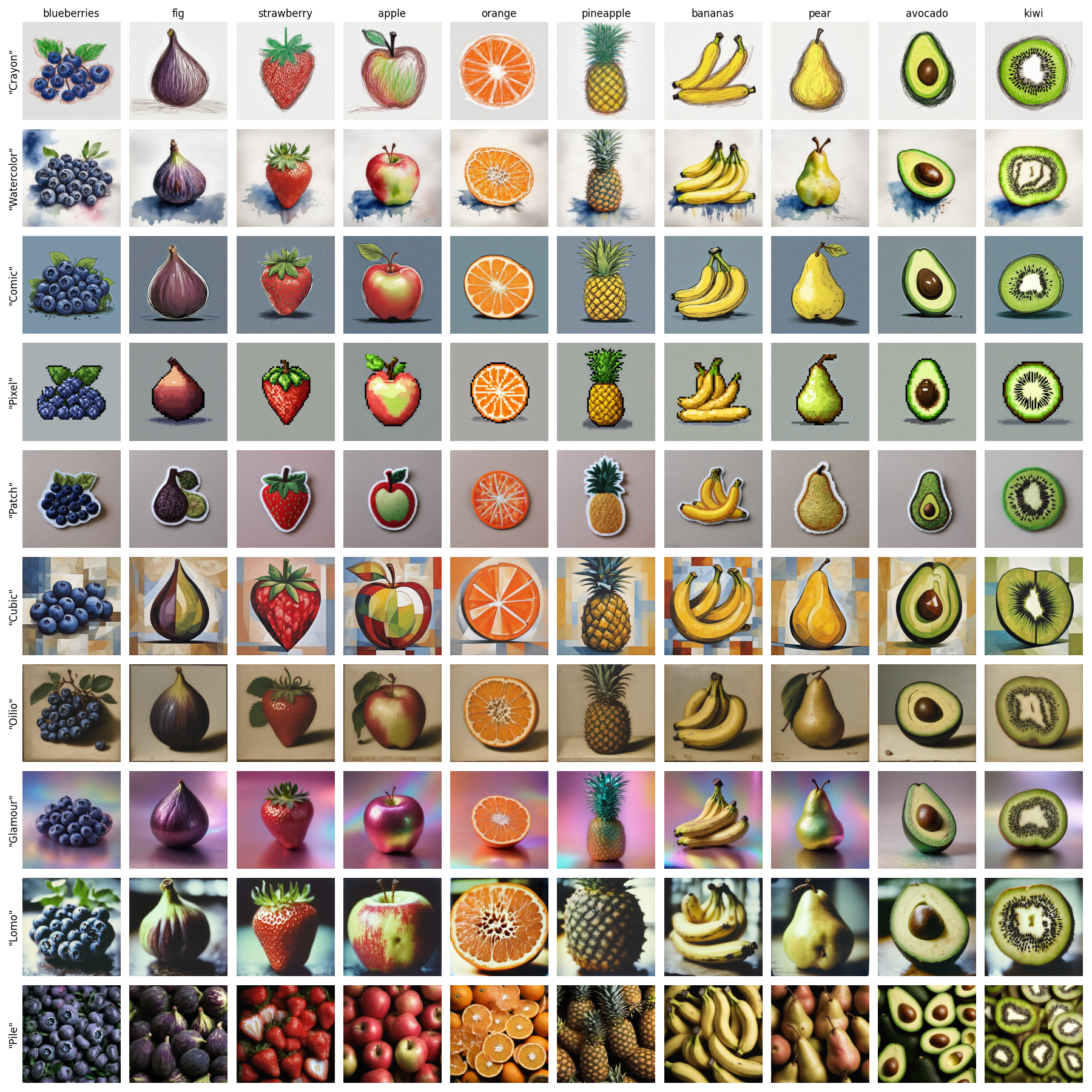}
\caption{\textit{Overview of one instance of 10 fruit categories in 10 styles. The columns display fruit categories including from left to right: blueberries, fig, strawberry, apple, orange, pineapple, bananas, pear, avocado, kiwi. The rows display style labels trying to describe the style prompts from top to bottom: 'Crayon', 'Watercolor', 'Comic', 'Pixel', 'Patch', 'Cubic', 'Oilio', 'Glamour', 'Lomo', 'Pile'. The full dataset contains 100 instances for each fruit category-style combination.}}
\label{fig:overview}
\end{figure}

\section*{Background \& Summary}
Similarity perception is an abstract and complex concept that differs widely across mental and computational models, as explored in (computational) neuroscience \cite{kaiser2022Modelling, charest2014Unique}, computer vision \cite{lang2018Attesting, wei2022Difference, muttenthaler2024improving,fu2023DreamSim}, or (computational) cognitive science \cite{hummel2023Analogy, richie2021Similarity}. For mental and conceptual models, similarity refers to resemblance or alikeness and describes groups with some shared properties, as prominently outlined in Wittgenstein's remarks on family resemblance \cite{wittgenstein1968Philosophical, rosch1975Familya}. Conversely, in computational models, similarity denotes proximity and is conventionally defined as inversely correlated with distance between data points in a metric space. 

Computer Vision applications heavily rely on such visual similarity, often utilizing vector embeddings that set up a measurable multidimensional space to index images. In similarity learning the goal is to train models that can accurately capture the underlying similarities between data points, enabling tasks such as image retrieval or classification based on similarity metrics \cite{mishra2021Effectively,liu2019Neural,cheng2018Deep,veit2017Conditional,mathisen2020Learning,ohsong2016Deep,ma1996Texture}. However, similarity in these contexts is often implied to be understood in a singular notion, overlooking the multifaceted nature of similarity perception crucial for informed decision-making in selecting models or methods. For instance, Ref \cite{ruiz2023DreamBooth} utilizes CLIP \cite{radford2021Learning} and DINO \cite{caron2021Emerging} to evaluate subject fidelity of generated images, acknowledging the varying importance of different similarity aspects. It is generally considered that CLIP captures semantic relationships, while DINO focuses more on visual features. Yet, validating such assumptions poses a significant challenge. 

Research in quantitative and computational aesthetics \cite{karjus2023Compression, forsythe2011predicting, zhang2020inkthetics}, as well as the interplay of computation and human cultures \cite{brinkmann2023machine, mccormack2022complexity}, requires reliable benchmark datasets that are interpretable by machines and humans. Previous work has relied on embeddings of large amounts of well known artworks \cite{srinivasa2022wikiartvectors, mao2017deepart} or synthetic datasets of limited size \cite{PhysRevResearch, ostmeyer2024synthetic, dunabeitia2018multipic, ovalle2022standardized}.

Benchmark image datasets for perceptual similarity judgment exist, with some relying on annotated text captions of real-world images \cite{liaoArtBench}, while others utilize synthetic image triplets designed to better align with mental models \cite{fu2023DreamSim}. However, these datasets primarily focus on specific tasks or aspects of similarity perception and alignment, such as zero-shot evaluation or similarity metric optimization.

Here we propose Style Aligned Artwork Datasets (SALADs), with the fruit-SALAD serving as an exemplar. This synthetic image dataset comprises 10,000 generated images featuring 10 easily recognizable fruit categories, each represented in 10 visually distinct styles, with 100 instances each. The deliberate control over semantic and stylistic properties inherent to each image facilitates comparative analysis of different image embedding and complexity models, enabling an exploration of their similarity perception.

We characterize the dataset through various machine learning models and complexity measures, showcasing how simple pairwise comparisons of image vectors can yield robust inter-comparable measures. Our examples reveal significant differences in similarity awareness across these methods and models, shedding light on anecdotal considerations stemming from differences in model or algorithm design, training data, parameter configuration, or similarity measures. In turn, this approach can be used to guide model training and alignment.

The fruit-SALAD offers opportunities for joint robust quantification and qualitative human interpretation, enhancing algorithmic and human perception regarding differences in measuring vector similarity and visual resemblance across computational, and statistical models. This approach allows for a more comprehensive assessment of similarity perception, beyond the scope of existing benchmark datasets, ultimately contributing to a deeper understanding of computational and human similarity perception mechanisms.

\section*{Methods}
\subsection*{Image Generation}
We used Stable Diffusion XL (SDXL)\cite{podell2023SDXL} and StyleAligned \cite{hertz2024Style} to create the fruit-SALAD by carefully crafting image generation prompts and supervising the automation process. Diffusion probabilistic models \cite{ho2020Denoising} are typically trained with the objective of denoising blurred images. By leveraging their ability to iteratively refine images by processing random noise, these models can be used in conjunction with text prompts to generate images. Such Text-to-Image models, exemplified by DALL-E \cite{ramesh2021ZeroShot, openaiDALLE}, Midjourney \cite{midjourney} and StableDiffusion \cite{rombach2022HighResolutiona}, have recently gained significant attention in various creative and commercial domains. These models and services have simplified the synthesis of high-quality individual images, enabling unprecedented ease of use through natural language. However, scaling the generation process or achieving stylistically consistent images remains challenging but can be improved by style alignment method \cite{hertz2024Style} to coordinate shared attention across multiple generations based on a reference style image. 

\begin{figure}[hbt!]
\centering
\includegraphics[width=\linewidth]{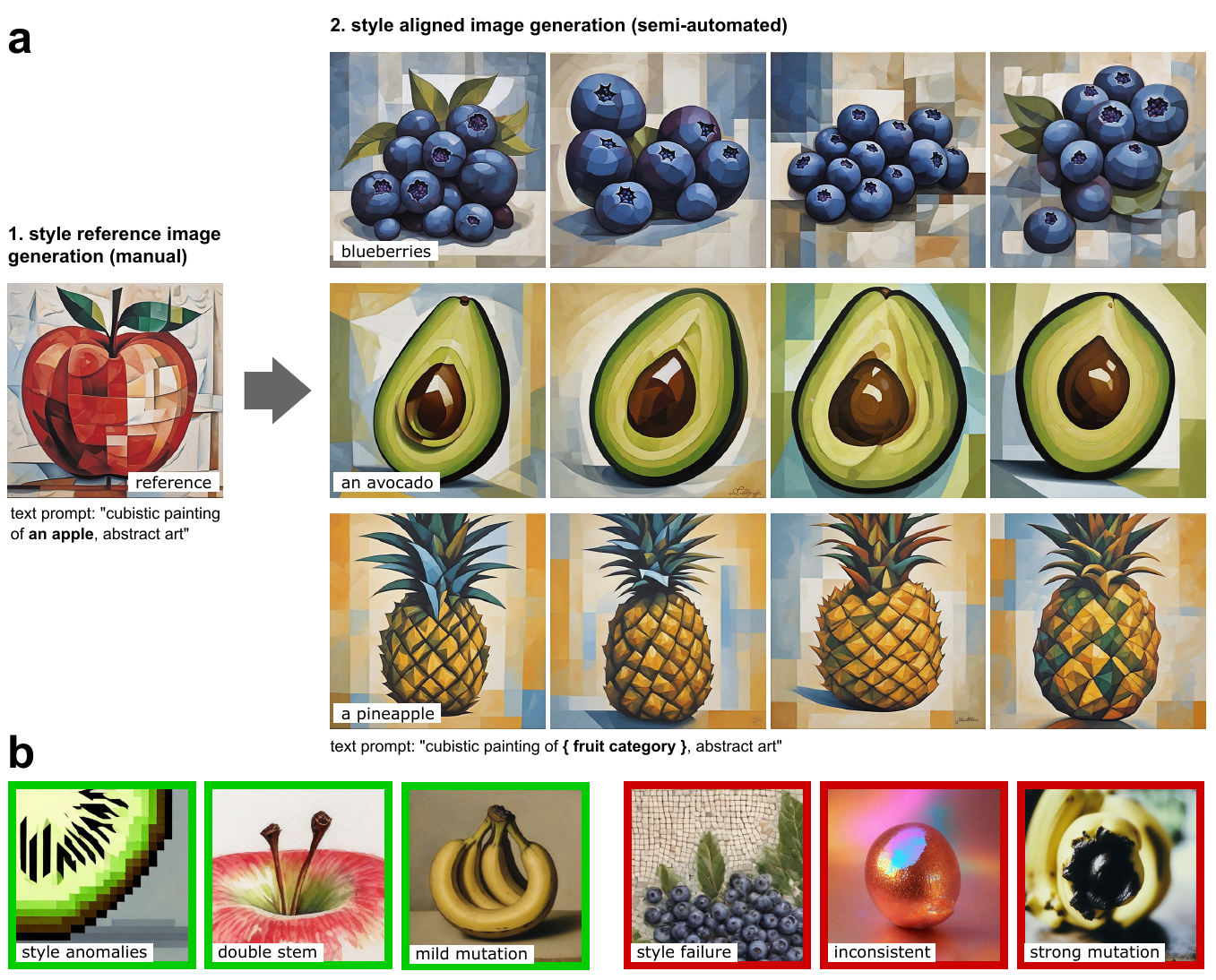}
\caption{\textit{Overview of the image generation process. (a) Image generation pipeline. 1. Style reference image generation with Stable Diffusion XL \cite{podell2023SDXL} in manual trial-and-error fashion using text prompts of style description in combination with ‘an apple’. 2. Style aligned image generation \cite{hertz2024Style} based on each style reference image using diffusion inversion and text prompts iterating over 10 fruit categories generating 100 instances each, resulting in 10,000 images. (b) Examples of tolerated and rejected results. Left: tolerated minor issues which do not impact recognition of category or style; right: rejected major issues which are either unrecognizable or inconsistent across the style. }}
\label{fig:pipeline}
\end{figure}

We utilized a computational approach to scale the image generation process. Initially, we experimented in a trial-and-error fashion with different style prompts in conjunction with different fruit categories, using SDXL \cite{podell2023SDXL} for image generation. Successful results were selected as style references. We then used style alignment \cite{hertz2024Style} to generate multiple instances of different fruits within the same style using diffusion inversion \cite{song2022denoising} of the reference image. Through several iterations and adjustments to the prompts, we refined the process and eventually automated the generation to produce 100 instances for each fruit-style combination.

The fruit prompts and stylistic references we selected were carefully curated to improve the robustness of the style alignment generation method. Among the fruit prompts, we balanced between fruit prototypicality and variability across different stylistic prompts to ensure compatibility with generation on scale, while simultaneously covering a wide range of fruit shapes and colors. Similarly, our selection of stylistic references was based on their effectiveness in aligning with the generation space, focusing on those that demonstrated superior performance in achieving stylistic coherence (see examples in Fig. \ref{fig:overview} and Fig. \ref{fig:pipeline}b).

We maintained dataset quality by visually assessing the entire dataset in 100 batches of 10 by 10 image grids (see example in Fig. \ref{fig:overview}) and manually replaced images that were inconsistent across all instances. Therefore, the final dataset with category and style classes may be biased by our own aesthetic arbitration, which is akin to the inherent specificity of a chosen set of handwritten digits \cite{lecun1998Gradientbased}.

\subsection*{Image Embeddings}
Our exemplary vector embeddings are derived from machine learning models and compression algorithms through various commonly employed methods (Table \ref{tab:pretrained}). For \cite{dosovitskiy2021Imageb, caron2021Emerging, oquab2024DINOv2, radford2021Learning, woo2023ConvNeXt, liu2022ConvNet} we extracted feature vectors using the flattened last hidden states. For \cite{he2015Deep, simonyan2015Very, chollet2017Xception} we used average pooling from the second to last layer.

\begin{table}[h!]
\centering
\begin{tabular}{|l|l|l|l|}
\hline
\textbf{model short name}& \textbf{type}& \textbf{training set}&\textbf{dimensions}\\
\hline
ViT-B-16\textunderscore{}IN21k& Vision Transformer \cite{dosovitskiy2021Imageb} (base, 16x16)&  ImageNet-21k \cite{deng2009ImageNet}&768\\
\hline
ViT-B-32\textunderscore{}IN21k& Vision Transformer \cite{dosovitskiy2021Imageb} (base, 32x32)&  ImageNet-21k \cite{deng2009ImageNet}&768\\
\hline
 ViT-H-14\textunderscore{}IN21k& Vision Transformer \cite{dosovitskiy2021Imageb} (huge, 14x14)& ImageNet-21k \cite{deng2009ImageNet}&1280\\\hline
 DINO\textunderscore{}IN1k& DINO \cite{caron2021Emerging}, Vision Transformer \cite{dosovitskiy2021Imageb} (base, 16x16)& ImageNet-1k \cite{deng2009ImageNet}&768\\\hline
 DINOv2\textunderscore{}B\textunderscore{}LVD& DINOv2 \cite{oquab2024DINOv2} (base)&  LVD-142M \cite{oquab2024DINOv2}&768\\\hline
 ResNet50\textunderscore{}IN1k& ResNet \cite{he2015Deep}& ImageNet-1k \cite{deng2009ImageNet}&2,048\\\hline
 VGG19\textunderscore{}IN1k& VGG \cite{simonyan2015Very}& ImageNet-1k \cite{deng2009ImageNet}&512\\\hline
 Xception\textunderscore{}IN1k& Xception \cite{chollet2017Xception}& ImageNet-1k \cite{deng2009ImageNet}&2,048\\\hline
 ConvNeXt\textunderscore{}L400M& ConvNeXt \cite{liu2022ConvNet} (base)& LAION-400M \cite{schuhmann2021LAION400Ma}&512\\\hline
 ConvNeXt-v2\textunderscore{}L400M& ConvNeXt-V2 \cite{woo2023ConvNeXt} & LAION-400M \cite{schuhmann2021LAION400Ma}&320\\\hline
 CLIP-ViT-B-16\textunderscore{}L2B& CLIP \cite{radford2021Learning}, Vision Transformer \cite{dosovitskiy2021Imageb} (base, 16x16)& LAION-2B \cite{schuhmann2022LAION5B}&512\\\hline
 CLIP-ViT-B-32\textunderscore{}L2B& CLIP \cite{radford2021Learning}, Vision Transformer \cite{dosovitskiy2021Imageb} (base, 32x32)& LAION-2B \cite{schuhmann2022LAION5B}&512\\\hline
 CLIP-ViT-H-14\textunderscore{}L2B& CLIP \cite{radford2021Learning}, Vision Transformer \cite{dosovitskiy2021Imageb} (huge, 14x14)& LAION-2B \cite{schuhmann2022LAION5B}&1,024\\\hline
 CLIP-ViT-B-16\textunderscore{}L400M& CLIP \cite{radford2021Learning}, Vision Transformer \cite{dosovitskiy2021Imageb} (base, 16x16)& LAION-400M \cite{schuhmann2021LAION400Ma}&512\\\hline
 CLIP-ViT-B-16\textunderscore{}OA& CLIP \cite{radford2021Learning}, Vision Transformer \cite{dosovitskiy2021Imageb} (base, 16x16)& OpenAI (undisclosed)&512\\\hline
 CLIP-RN50\textunderscore{}OA& CLIP \cite{radford2021Learning}, ResNet50 \cite{he2015Deep}& OpenAI (undisclosed)&1,024\\\hline
 CLIP-RN101\textunderscore{}OA& CLIP \cite{radford2021Learning}, ResNet101 \cite{he2015Deep}& OpenAI (undisclosed)&512\\\hline
\end{tabular}
\caption{\label{tab:pretrained} Pre-trained Machine Learning models used for feature extraction.}
\end{table}

\begin{table}[h!]
\centering
\begin{tabular}{|l|l|l|}
\hline
\textbf{model short name}& \textbf{method}&\textbf{dimensions}\\
\hline
CompressionEnsembles& Compression Ensembles \cite{karjus2023Compression}&87\\
\hline
GIF\textunderscore{}compression& LZW \cite{welch1984Technique} to PNG file size ratios&1\\
\hline
 PNG\textunderscore{}filesizes& original PNG file sizes&1\\\hline
 style\textunderscore{}blind& one-hot encoding of fruit category only, ignoring styles&10\\\hline
 category\textunderscore{}blind& one-hot encoding of styles only, ignoring fruit category&10\\\hline
 balanced& one-hot encoding of fruit category and styles&20\\\hline
\end{tabular}
\caption{\label{tab:other}Other methods used for feature extraction.}
\end{table}

As an example of a quantitative aesthetics measure, we used the Compression Ensembles method \cite{karjus2023Compression}, which captures polymorphic family resemblance via a number of transformations (87 in our implementation). We used GIF image compression ratios, taking advantage of the Lempel–Ziv–Welch algorithm \cite{welch1984Technique}. We also provide the PNG file sizes as comparison (Table \ref{tab:other}).

To provide simple conceptual models for reference, we used binary, one-hot encoded vectors. In this encoding scheme, each vector represents a fruit category or style, with a value of \texttt{1} indicating the presence and \texttt{0} indicating the absence of the corresponding category or style (Table \ref{tab:other}). We are consciously providing a simple conceptual reference, to avoid the complications of full blown conceptual reference models, such as the CIDOC-CRM \cite{doerr2003cidoc}.

\section*{Technical Validation}


\subsection*{Self Recognition Test}
\begin{figure}[h!]
\centering
\includegraphics[width=\linewidth]{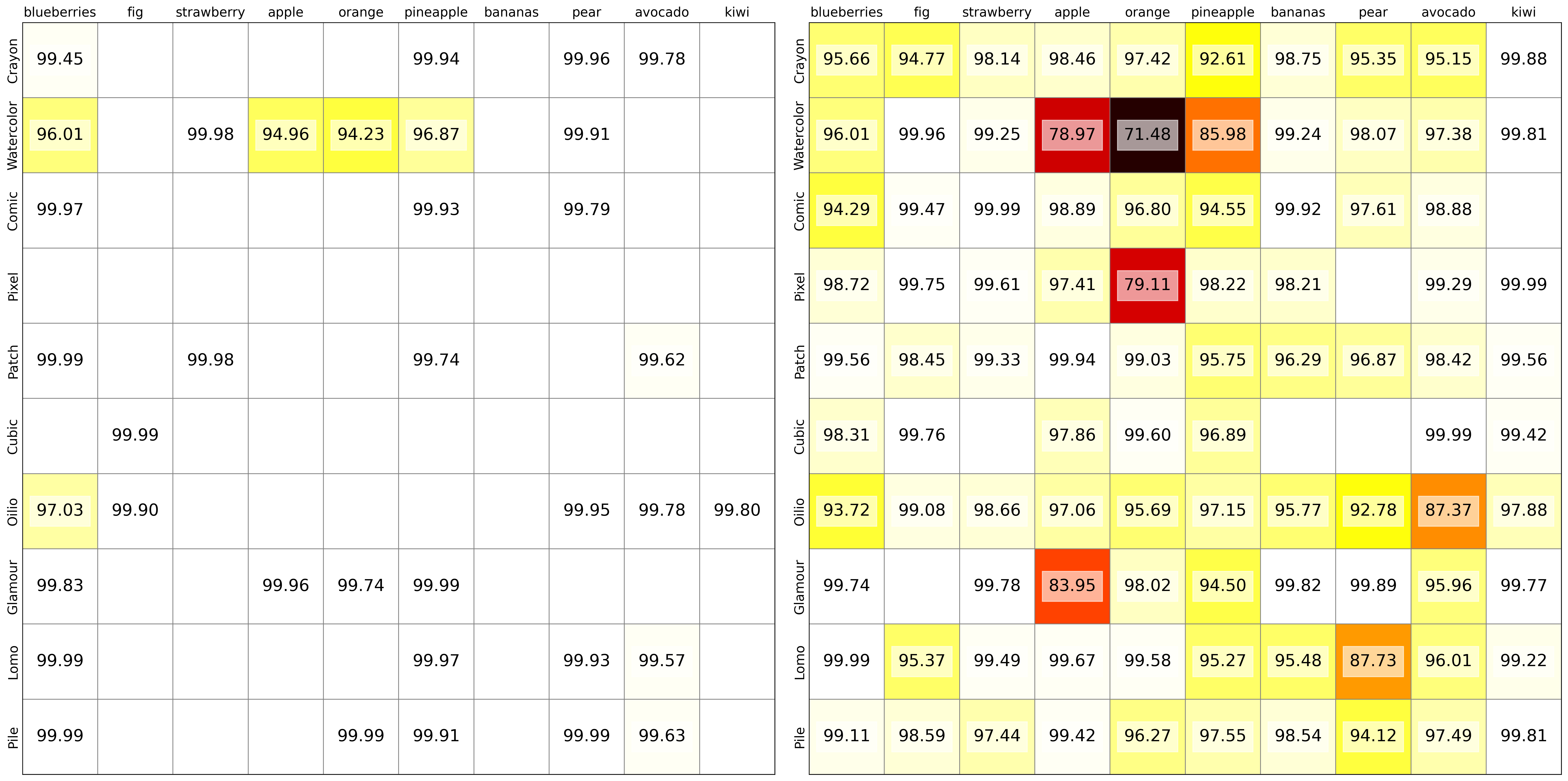}
\caption{\textit{Self-recognition Tests. Each cell represents the mean number of same instances in the top 100 nearest neighbors of its fruit category (column) and style (row) combination images. White cells without values have a perfect score of 100 out of 100 correctly recognized instances. Left: Maximum values from all computational models, taking into account that high scores within 100 out of 10,000 images reflect higher than chance results. Right: ResNet50\textunderscore{}IN21k as an example model.}}
\label{fig:recognition}
\end{figure}

One expects that, despite inevitable variation in similarity perception, the similarity of images from the same category-style combination should be systematically larger than between images of different categories and/or styles. To assess this, we conduct a self-recognition test on the fruit-SALAD\textunderscore{}10k dataset. This test involves retrieving the top 100 Nearest Neighbors for each image and counting how many instances of the same category-style combination are found within this set. The average number of successful retrievals across all 100 instances per model is then calculated. To validate the self-recognition of image instances, we select the maximum values across all computational models (Fig. \ref{fig:recognition}). 

If a category-style combination cannot be sufficiently recognized in any of the computational models, we consider the self-recognition test failed. Notably, we found that "apples" and "oranges" in the "Watercolor" style pose the greatest challenge, achieving sufficient accuracy only after various iterations of image generation.

\subsection*{Model Heatmaps}

\begin{figure}[h!]
\centering
\includegraphics[width=\linewidth]{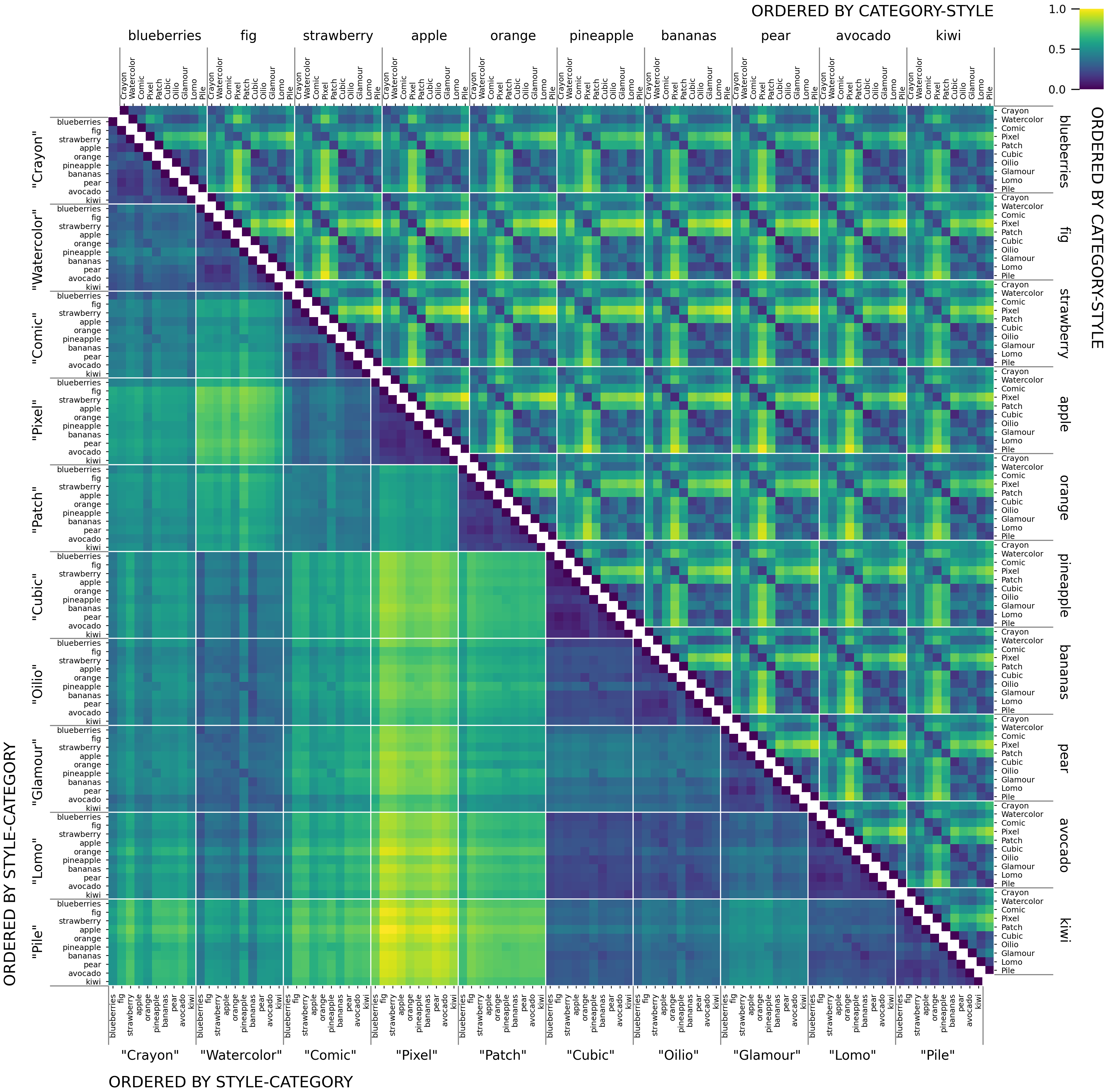}
\caption{\textit{DINO-ViT-B-16\textunderscore{}IN1k Heatmap indicating the mutual Mahalanobis distances of fruit-SALAD images. The matrix cells correspond to the mean of all 10,000 distance pairs of 100 by 100 instances of fruit-SALAD images. Below the diagonal: sorted by style first and fruit category second. Above the diagonal: sorted by fruit category first and style second. The color indicates the pairwise Mahalanobis distance of image embedding vectors obtained from the respective model or algorithm, from low to high (blue to yellow) while low values indicate higher similarity. The figure construction is comprehensive as the matrices are symmetric; diagonal cells can be left out.}}
\label{fig:dino-heatmaps}
\end{figure}

We characterize the dataset, and concurrently exemplify its possible future use by a set of category- and style-ordered distance matrices, which demonstrate salient differences in category and style similarity weights, across various computational models (Fig. \ref{fig:dino-heatmaps}). As a measure of similarity between two sets of images we calculate the average distances between all pairs of elements. To better generalize standardization, we use Mahalanobis distance\cite{2018Reprint, demaesschalck2000Mahalanobis}, which normalizes and decorrelates the coordinates.

\subsection*{Model Comparison}
\begin{figure}[h!]
\centering
\includegraphics[width=\linewidth]{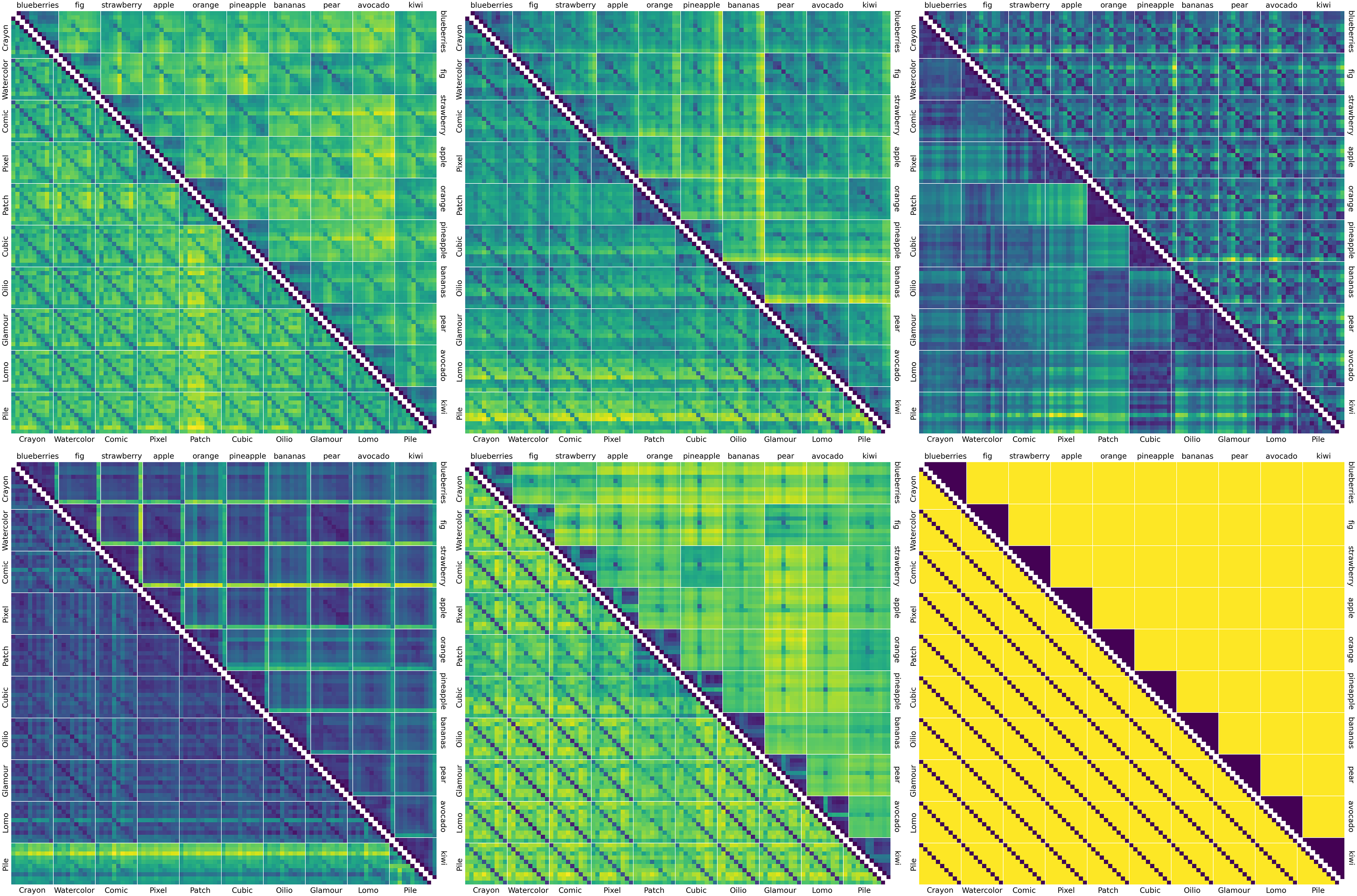}
\caption{\textit{Heatmaps indicating the mutual Mahalanobis distance of fruit-SALAD images according to different models (see Fig. \ref{fig:dino-heatmaps}). Top row from left to right: CLIP-ViT-B-16\textunderscore{}L400M, DINOv2-B\textunderscore{}LVD, CompressionEnsembles. Bottom row from left to right: VGG19\textunderscore{}IN1k, ViT-B-32\textunderscore{}IN21, style\textunderscore{}blind. The matrix ordering is identical.}}
\label{fig:multi-heatmaps}
\end{figure}

\begin{figure}
\centering
\includegraphics[width=\linewidth]{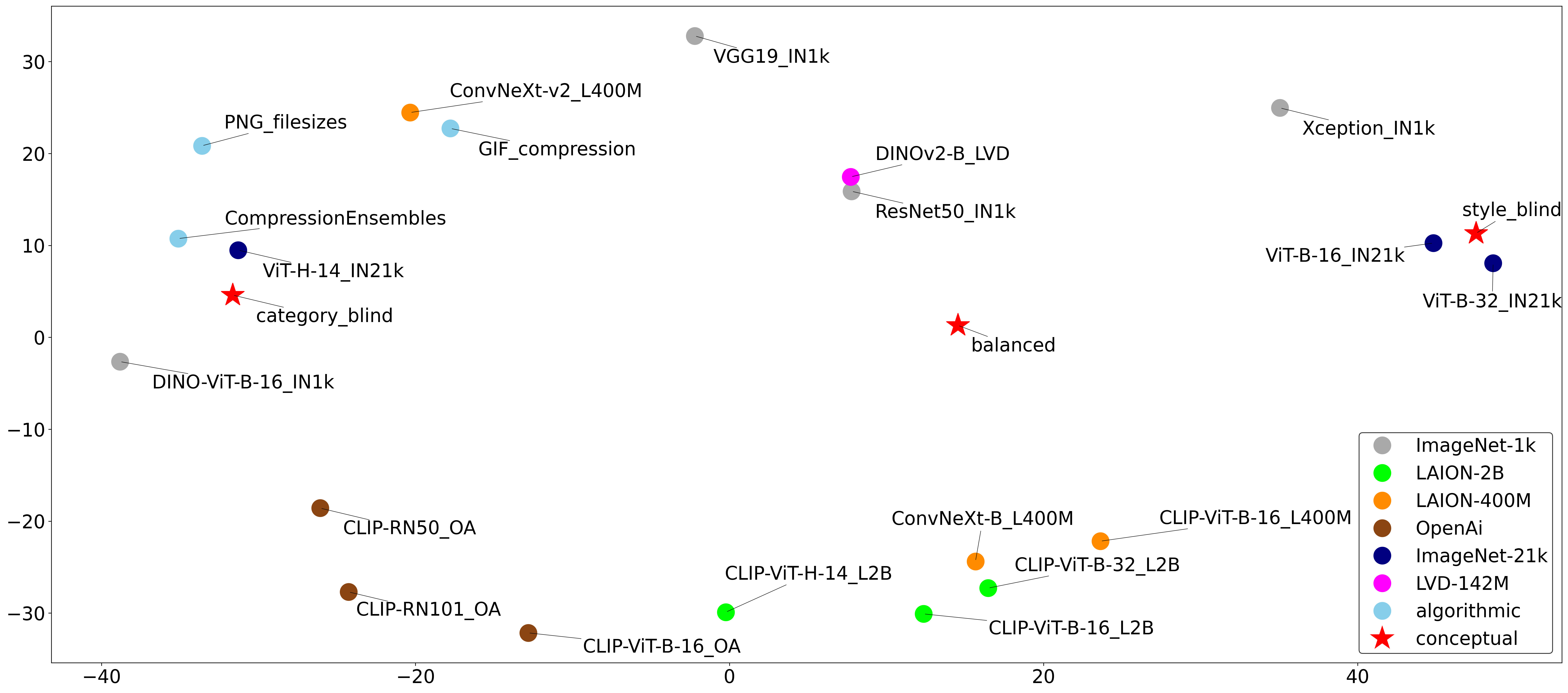}
\caption{\textit{Relative model comparison using principal component analysis (PCA) based on 23 standardized model vectors of 4,950 dimensions. These dimensions encompass the mutual Mahalanobis distances of all unique category-style combinations of the fruit-SALAD images, excluding self-pairing. Each fruit category-style combination is the mean of all 10,000 mutual distances of 100 by 100 fruit-SALAD image instances.
}}
\label{fig:model-comparison}
\end{figure}

Each of the multiple embedding models can be characterized by a set of distances between images in this embedding. One can consider this set of distances as a multidimensional vector, characterizing a model. Thus, the different models are represented as vectors in a shared space, which enables their direct comparison.
As coordinates we used standardized pairwise distances between all unique pairs of 100 fruit category-style combinations, i.e., all entries of the model heatmaps in Fig. \ref{fig:multi-heatmaps}. The principle components of the resulting embedding are shown in Fig. \ref{fig:model-comparison}.

\FloatBarrier

\begin{figure}[h!]
\centering
\includegraphics[width=\linewidth]{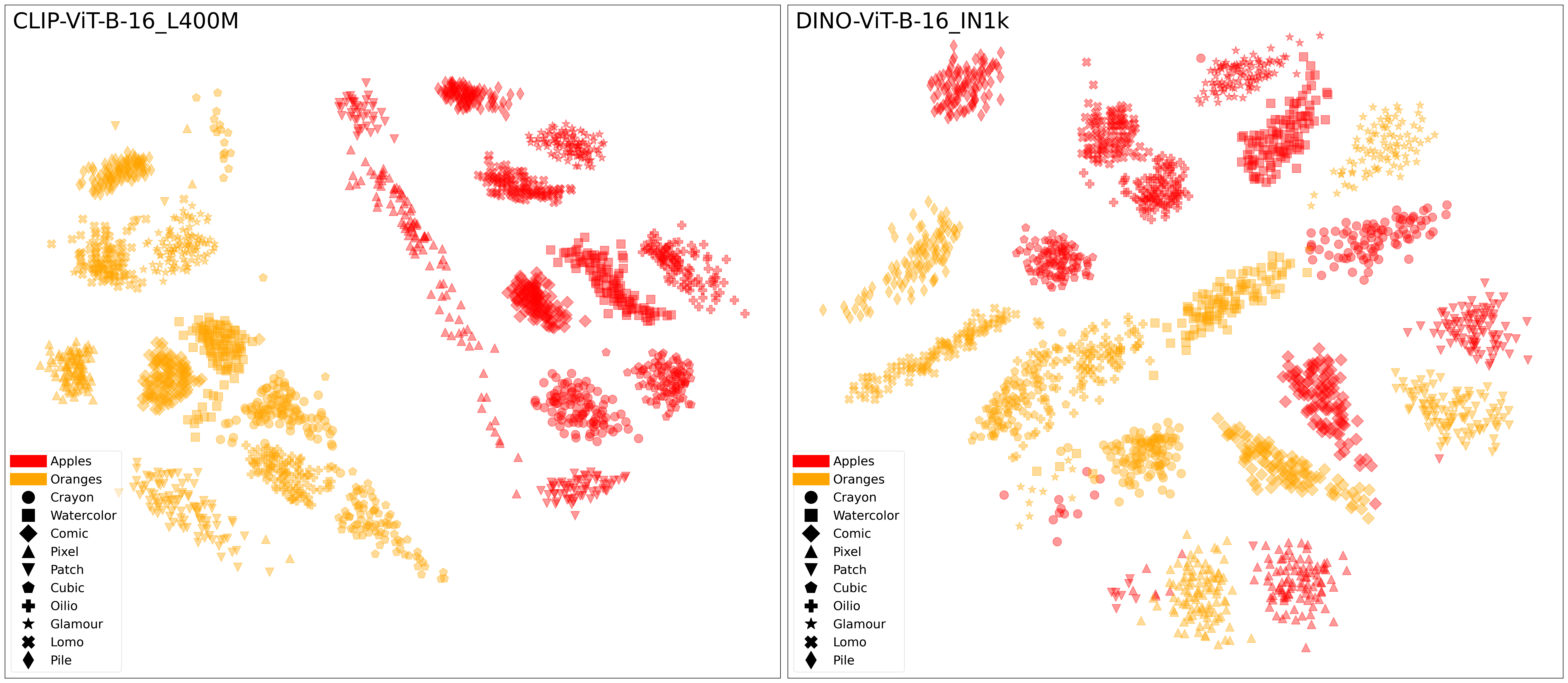}
\caption{\textit{Scatter plots of apples and oranges using multidimensional scaling (MDS) based on normalized image embedding vectors from two different models. Left: CLIP-ViT-B-16\textunderscore{}L400M; right: DINO-ViT-B-16\textunderscore{}IN1k. Colors indicate fruit categories and dot shapes indicate styles.}}
\label{fig:apples}
\end{figure}

Investigating the differences in similarity perception can also be accomplished by examining fruit categories and styles through the image embeddings of individual models (Fig. \ref{fig:apples}).

\FloatBarrier

\section*{Data Records \& Code Availability}
The fruit-SALAD consists of 10,000 PNG images of 1024x1024 pixels; 10 fruit categories in 10 styles with 100 instances each. 

The fruit-SALAD\textunderscore{}10k image dataset, style reference images, used text prompts, 100 grid overview plots (10x10 images per instance), vector files and model heatmaps are available at \href{https://doi.org/10.5281/zenodo.11158522}{https://doi.org/10.5281/zenodo.11158522}.

The image filenames adhere to the following format: \texttt{fruit\textunderscore{}style\textunderscore{}instance.png}. For example, an image with the filename \texttt{8\textunderscore{}1\textunderscore{}42.png} signifies fruit category \texttt{8} (avocado) rendered in style category \texttt{1} (Watercolor), and it represents generation number \texttt{42}. 

For accessibility, we provide all vector files as comma-separated values (.csv) with image file names as indices. 

The data repository also includes all style reference images, used text prompts, 100 grid overview plots (10x10 images per instance), and model heatmaps. 

Code performed to generate the fruit images is available at \href{https://github.com/tillmannohm/fruit-SALAD}{https://github.com/tillmannohm/fruit-SALAD}.

\section*{Acknowledgements}
All authors are supported by the CUDAN ERA Chair project for Cultural Data Analytics, funded through the European Union’s Horizon 2020 research and innovation program (Grant No. 810961).

\section*{Author contributions}
TO co-designed the research, generated the dataset, co-wrote the text, and created the figures. MT, AK and MS also co-designed the research, co-wrote the text, and provided conceptual guidance. All authors read and approved the final manuscript.

\bibliography{SALADs}

\end{document}